\documentclass{llncs}

\usepackage{graphicx}
\usepackage{enumitem}

\usepackage[figure]{algorithm2e}

\title{Distributed solving through model splitting}

\author{Lars Kotthof\/f \and Neil C.A.\ Moore\\
\email{\{larsko,ncam\}@cs.st-andrews.ac.uk}}
\institute{University of St Andrews}

\begin{document}

\maketitle

\begin{abstract}
Constraint problems can be trivially solved in parallel by exploring different
branches of the search tree concurrently. Previous approaches have focused on
implementing this functionality in the solver, more or less transparently to the
user. We propose a new approach, which modifies the constraint model of the
problem. An existing model is split into new models with added constraints
that partition the search space. Optionally, additional constraints are imposed
that rule out the search already done. The advantages of our approach are that
it can be implemented easily, computations can be stopped and restarted, moved
to different machines and indeed solved on machines which are not able to
communicate with each other at all.
\end{abstract}

\section{Introduction}

Constraint problems are typically solved by searching through the possible
assignments of values to variables. After each such assignment, propagation can
rule out possible future assignments based on past assignments and
the constraints. This process builds a search tree that explores the space of
possible (partial) solutions to the constraint problem.

There are two different ways to build up these search trees -- $n$-way branching
and $2$-way branching. This refers to the number of new branches which
are explored after each node. In $n$-way branching, all the $n$ possible
assignments to the next variable are branched on. In $2$-way branching, there
are two branches. The left branch is of the form $x=y$ where $x$ is a variable
and $y$ is a value from its domain. The right branch is of the form $x\neq y$.

The more commonly used way is $2$-way branching, implemented for example in
the Minion constraint
solver~\cite{minion}\footnote{\url{http://minion.sf.net}}. However, regardless
of the way the branching is done, exploring the branches can be done
concurrently. No information between the branches needs to be exchanged in order
to find a solution to the problem.

We exploit this fact by, given the model of a constraint problem, generating new
models which partition the remaining search space. These models can then be
solved independently. We furthermore represent the state of the search by adding
additional constraints such that the splitting of the model can occur at any
point during search. The new models can be resumed, taking advantage of both the
splitting of the search space and the search already performed.

\section{Background}

The parallelisation of depth-first search has been the subject of much
research in the past. The first papers on the subject study the distribution
over various specific hardware architectures and investigate how to achieve good
load balancing~\cite{rao_parallel_1987,kumar_parallel_1987}. Distributed solving
of constraint problems specifically was first explored only a few years
later~\cite{collin_feasibility_1991}.

Backtracking search in a distributed setting has also been investigated by
several authors~\cite{rao_efficiency_1993,sanders_better_1995}. A special
variant for distributed scenarios, asynchronous backtracking, was proposed
in~\cite{yokoo_distributed_1992}. Yokoo \textit{et
al} formalise the distributed constraint satisfaction problem and present
algorithms for solving it~\cite{yokoo_distributed_1998}.

Schulte presents the architecture of a system that uses
networked computers~\cite{schulte_parallel_2000}. The focus of his approach is
to provide a high-level and reusable design for parallel search and achieve a
good speedup compared to sequential solving rather than good resource
utilisation. More recent papers have explored how to transparently parallelise
search without having to modify existing code~\cite{michel_parallelizing_2007}.

Most of the existing work is concerned with the problem of effectively
distributing the workload such that every compute node is kept busy. The most
prevalent technique used to achieve this is work stealing. The compute nodes
communicate with each other and nodes which are idle request a part of the work
that a busy node is doing. Blumofe and Leiserson propose and discuss a work
stealing scheduler for multithreaded computations
in~\cite{blumofe_scheduling_1999}. Rolf and Kuchcinski investigate different
algorithms for load balancing and work stealing in the specific context of
distributed constraint solving~\cite{rolf_load-balancing_2008}.

Several frameworks for distributed constraint solving have been proposed and
implemented, e.g.\ FRODO~\cite{frodo}, DisChoco~\cite{dischoco} and
Disolver~\cite{disolver}.  All of these approaches have in common that the
systems to solve constraint problems are modified or augmented to support
distribution of parts of the problem across and communication between multiple
compute nodes. The constraint model of the problem remains unchanged however; no
special constructs have to be used to take advantage of distributed solving. All
parallelisation is handled in the respective solver. This does not preclude the
use of an entirely different model of the problem to be solved for the
distributed case in order to improve efficiency, but in general these solvers
are able to solve the same model both with a single executor and distributed
across several executors.

The decomposition of constraint problems into subproblems which can be solved
independently has been proposed in~\cite{michel_decomposition-based_2004},
albeit in a different context. In this work, we explore the use of this
technique for parallelisation. A similar approach was taken
in~\cite{rolf_load-balancing_2008}, but requires parallelisation support in the
solver.

\section{Model splitting}

We now describe our new approach to the distributed solving of constraint
problems which modifies the constraint solver to modify the constraint model and
does not require explicit parallelisation support in the solver.

Before splitting, the solver is stopped. As well as stopping, it is designed to
output \emph{restart nogoods} for the problem in the solver's own input language
\cite{restartnogood}.  These constraints, when added to the problem, will
prevent the search space just explored from being repeated in any split
model\footnote{This same technique allows Minion to be paused and resumed: the
  nogoods are provided when the solver is interrupted, and can be used to
  restart search, potentially using a different solver, different search
  strategy or on a different machine.}.

To split the search space for an existing model, partition the domain for the
variable currently under consideration into $n$ pieces of roughly equal size.
Then create $n$ new models and to each in turn add constraints ruling out $n-1$
partitions of that domain. Each one of these models restricts the possible
assignments to the current variable to one $n$th of its domain.

As an example, consider the case $n=2$. If the variable under consideration is
$x$ and its domain is $\{1,2,3,4\}$, we generate $2$ new models. One of them has
the constraint $x\leq 2$ added and the other one $x\geq 3$. Thus, solving the
first model will try the values $1$ and $2$ for $x$, whereas the second model
will try $3$ and $4$.

The main problem when splitting constraint problems into parts that can be
solved in parallel is that the size of the search space for each of the splits
is impossible to predict reliably. This directly affects the effectiveness of
the splitting however -- if the search space is distributed unevenly, some of
the workers will be idle while the others do most of the work.

We address this problem by providing the ability to split a constraint model
after search has started. The approach is very similar to the one explained
above. The only difference is that in addition to the constraints that partition
the search space, we also add constraints that rule out the search space that
has been explored already.

Assume for example that we are doing $2$-way branching, the variable currently
under consideration is again $x$ with domain $\{1,2,3,4\}$ and the branches that
we have taken to get to the point where we are are $x\neq 1$ and $x\neq 2$. The
generated new models will all have the constraints $x\neq 1$ and $x\neq 2$ to
get to the point in the search tree where we split the problem. Then we add
constraints to partition the search space based on the remaining values in the
domain of $x$ similar to the previous example.

Using this technique, we can create new chunks of work whenever a worker becomes
idle by simply asking one of the busy workers to stop and generate split models.
The search is then resumed from where it was stopped and the remaining search
space is explored in parallel by the two workers. Note that there is a runtime
overhead involved with stopping and resuming search because the constraints
which enable resumption must be propagated and the solver needs to explore a
small number of search nodes to get to the point where it was stopped before.
There is also a memory overhead because the additional constraints need to be
stored.

We have implemented this approach in a development version of Minion, which we
are planning to release to the public after further testing and verification.
Initial experiments showed that the overhead of stopping, splitting and resuming
is minimal and not significant for large problems.

In practice, we run Minion for a specified amount of time, then stop, split and
resume instead of splitting at the beginning and when workers become idle. The
algorithm is detailed in Figure~\ref{distalgo}. This creates an $n$-ary
split tree of models for $n$ new models generated at each split. Initially, the
potential for distribution is small but grows exponentially as more and more
search is performed.

\setlength{\algomargin}{0pt}
\SetAlFnt{\normalsize}
\begin{procedure}
\SetKwFunction{solved}{solved?}
\SetKwFunction{exhausted}{search space exhausted?}
\SetKwInOut{Input}{Input}
\SetKwInOut{Output}{Output}
\SetKwFor{For}{for}{do in parallel}{end}

\Input{constraint problem $X$, allotted time $T_{max}$ and splitting factor
$n\geq 2$}
\Output{a solution to $X$ or nothing if no solution has been found}
\BlankLine
run Minion with input $X$ until termination or $T_{max}$\;
\BlankLine
\uIf{\solved{$X$}}{
   terminate workers\;
   \Return{solution}\;
}
\uElseIf{\exhausted}{
    \Return{}\;
}
\Else{
   $X' \leftarrow X$ with new constraints ruling out search already
   performed\;
   split $X'$ into $n$ parts $X'_1, \ldots, X'_n$\;
   \BlankLine
   \For{$i \leftarrow 1$ \KwTo $n$}{
       distSolve($X'_n$, $T_{max}$,$n$)\;
   }
}
\caption{distSolve($X$,$T_{max}$,$n$): Recursive procedure to find the first
solution to a constraint problem distributed across several workers.}
\label{distalgo}
\end{procedure}

\section{Comparison to existing approaches}

We see the main advantage of our approach in not requiring any involved changes
to the constraint solving system to support distributed solving; in particular
communications between workers. Conventionally, distribution is achieved with
the aid of recomputation and cloning; established techniques used
e.g.\ in~\cite{schulte_parallel_2000}. We require two features of our solver:
partitioning using constraints, and ability to output restart nogoods. Our
system makes use of \emph{cloning}, which we call ``splitting'' and implement by
means of nogoods added to the constraint model in order to partition the domain
of a variable. However, where other systems use \emph{recompution}, our system
uses restart nogoods. In a system based on recomputation the clone begins at
specific search path, e.g.\ stolen from another worker; with restart nogoods
notionally multiple search paths are provided and the solver may explore these
in any way it wishes, not necessarily one after the other. It is merely a
convenient and compact way of encoding the situation where a solver is
relinquishing \emph{all} its remaining work.

Contemporary constraint solvers make it easy to change or amend the search
procedure to support distribution across several executors, but even then
changes to the constraint solving system are required. While an initial
implementation of distributed search can be done relatively quickly, handling
failure properly and supporting things like nodes being added and removed
dynamically requires significantly more effort. Our approach separates this part
completely from the constraint solving system.

There are several advantages to implementing distributed solving the way
given in Figure~\ref{distalgo}. First, by creating regular ``snapshots'' of
the search done, the resilience against failure increases. Every time we stop,
split and resume, the modified models are saved. As they contain constraints
that rule out the search already done, we only lose the work done after that
point if a worker fails. This means that the maximum amount of work we lose in
case of a total failure of all workers is the allotted time $T_{max}$ times the
number of workers $|w|$.

The fact that the modified models can be stored can also be exploited to move
the solving process to a different set of workers after it has been started
without losing any work. It furthermore means that we require no communication
between the individual workers solving the problem; they only need to be able to
receive the problem to solve and send the solution or split models back.

Another advantage is that small problems which Minion can solve within
the allotted time are not split and no distribution overhead is incurred.
Solving proceeds as it would in a standard, non-distributed fashion.

Our approach is particularly suitable for use with existing
grid-computing software or workload management systems such as
Condor\footnote{\url{http://www.cs.wisc.edu/condor/}}. Every time new models are
generated, they are submitted to the system which queues them and allocates a
worker as soon as one becomes available. By leveraging existing software to
perform this task, a huge amount of development time is saved and errors
are avoided. For large problems, the number of queued jobs will usually exceed
the number of workers, ensuring good resource utilisation.

The management system to monitor the search, queue split models and terminate
the workers if a solution has been found can be implemented efficiently in just
a few lines of code. We have written a Ruby script that performs this task in
little more than an hour. Obviously there is potential for trying different
search strategies for different branches or modifying other search parameters in
order to improve efficiency. With the appropriate modifications, the management
system could adapt the search procedure specifically for individual parts of the
search tree. We are planning to explore these possibilities in future work.

A downside of the approach is that the number of models which can be solved in
parallel will be small to start with. This means that the utilisation of
resources in the beginning will be suboptimal. Only as more and more search
space is explored and more and more split models are generated, the utilisation
will improve. This however can be mitigated by dynamically adapting the time for
which Minion is run before splitting the problem -- in the beginning, we set
it to a small value to quickly get many models that we can solve in parallel.
Then we gradually increase the allotted time as the resource utilisation
improves.

Our technique is intended to be used for very large problems which take a long
time (many hours, days or weeks) to solve. It is unlikely to be efficient for
problems that can be solved in minutes, but on the other hand there is no need
for distributed solving if the problem can be solved sequentially in a short
amount of time. Only large search spaces can be split in a way that many workers
are kept busy without a high communication overhead.

\section{Detailed example}

We will now have a detailed look at how our approach works for a specific
problem. Consider the $4$-queens problem. We want to place $4$ queens on a
$4\times 4$ chessboard such that no queen is attacking another queen. Queens can
move along rows, columns and diagonals. The constraints therefore have to forbid
that two or more queens are in the same row, the same column or on the same
diagonal. The constraint model in Figure~\ref{queens} captures this problem.

\begin{figure}[htb]
{\normalsize
\begin{verbatim}
language Dominion 0.1
letting n = 4
dim queens[n]: int
find queens[..]: int {1..n}
such that
alldifferent alldiff(queens[..])
diagonals1 [ not(eq1 eq(queens[i], add(queens[j], j-i))) |
    i in {0..n-2}, j in {i+1..n-1} ]
diagonals2 [ not(eq2 eq(queens[i], add(queens[j], i-j))) |
    i in {0..n-2}, j in {i+1..n-1} ]
\end{verbatim}
\caption{Model for the $4$-queens problem in the Dominion
language~\cite{domlang}. The model describes the $n$-queens problem in general
and is specialised for $4$-queens in the second line.}
\label{queens}}
\end{figure}

\begin{figure}[htb]
{\normalsize
\includegraphics[width=\textwidth]{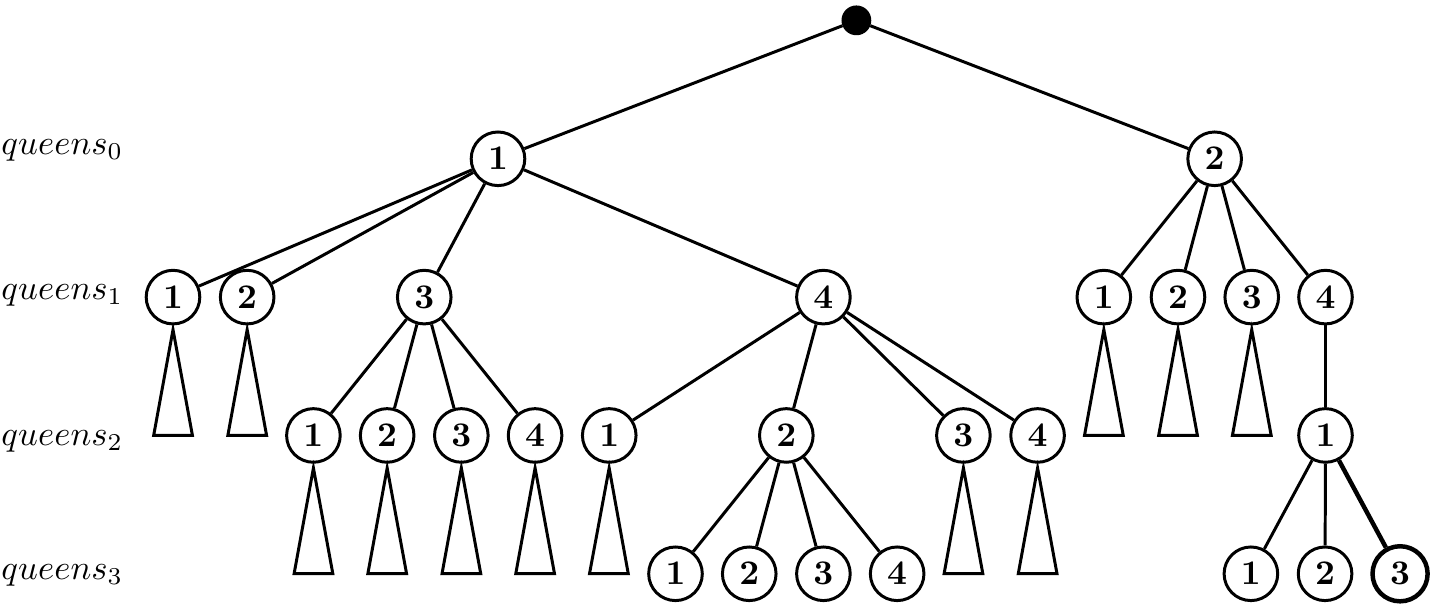}
\caption{First solution search tree for $4$-queens. The triangles depict
subtrees which are not explored because the partial assignment so far cannot be
part of a solution. The bold, rightmost node is where the solution is found. The
levels of the tree show assignments to the variables shown on the left.}
\label{searchtree}}
\end{figure}

We assume variable ordering $queens_0, queens_1, queens_2, queens_3$, ascending
value ordering from $1$ to $4$ and $n$-way branching. The search tree for a
simple backtracking algorithm is depicted in Figure~\ref{searchtree}. Even for a
very small problem like this, there is significant potential for distributed
solving.

We now start solving the problem until we reach the assignment $queens_0 = 2$.
Then we stop. The constraint we need to add to resume the search at the same
point is
\begin{center}
\verb+resume not(innerresume eq(queens[0], 1))+
\end{center}
(note that \texttt{resume} and \texttt{innerresume} are simply identifiers given
to the specific constraints as required in the Dominion
language~\cite{domlang}).

Let us assume a splitting factor of $2$. We add the constraints to split
the remaining search space as follows. The variable currently under
consideration is $queens_1$, its domain is $\{1,2,3,4\}$ and therefore the
constraints are
\begin{center}
\verb+left leq(queens[1], 2)+ \hspace*{2em} and \hspace*{2em} \verb+right leq(3, queens[1])+.
\end{center}

The search is restarted with two workers, each exploring separate branches
of the remaining search space. The first worker finds no
solutions in its part of the search space, terminates and returns. The second
worker finds a solution and returns it. Search terminates and no further
splitting is performed.

\section{Conclusions and future work}

We have proposed and detailed a novel approach for distributing constraint
problems across multiple computers. Instead of modifying the solver to support
distributed operation, we only require some simple and generic modifications
that post additional constraints to the model.

The main advantages of our approach are that it does not require networked
machines, is resilient against failure and can be implemented easily in
constraint solvers which are aware of the state of the search.

The main drawback of this paper is that we do not have performed a systematic
experimental evaluation of our approach yet. In the future, we would like to
evaluate it in terms of solving speedup and resource utilisation on large,
real-world problems. Furthermore, we would like to investigate finding all
solutions to a constraint problem and solving constrained optimisation problems
in a distributed manner.

Adapting the search procedure and parameters dynamically during search is
another promising area for future work. The solving process could be tailored
to the characteristics of parts of the search space to improve efficiency.

Another direction for future work is to support a higher level of abstraction
for decomposing problems into subproblems. This would be necessary to support
problems which cannot be decomposed by simply adding constraints that split the
domain of a variable.

\section*{Acknowledgements}

The authors thank Chris Jef\/ferson for help with implementing the model
splitting in Minion and the anonymous reviewers for their feedback. Lars
Kotthof\/f is supported by a SICSA studentship.

\bibliography{split}
\bibliographystyle{splncs03}

\end{document}